\documentclass[manuscript,screen]{acmart}
\usepackage{multirow}
\usepackage{adjustbox}
\AtBeginDocument{%
  \providecommand\BibTeX{{%
    \normalfont B\kern-0.5em{\scshape i\kern-0.25em b}\kern-0.8em\TeX}}}

\setcopyright{acmcopyright}
\copyrightyear{2018}
\acmYear{2018}
\acmDOI{10.1145/1122445.1122456}

\acmConference[Review]{review}{NY}
\acmBooktitle{}
\acmPrice{15.00}
\acmISBN{978-1-4503-XXXX-X/18/06}

\usepackage{ulem}

\begin{document}

\title{Semi-Supervised Learning for Eye Image Segmentation}


\author{Aayush K. Chaudhary}
\authornote{Both authors contributed equally to this research.}
\email{akc5959@rit.edu}
\orcid{1234-5678-9012}
\author{Prashnna K. Gyawali}
\authornotemark[1]
\email{pkg2182@rit.edu}
\affiliation{%
  \institution{Rochester Institute of Technology}
  \streetaddress{54 Lomb Memorial Drive}
  \city{Rochester}
  \state{New York}
  \country{USA}
  \postcode{14623}
}

\author{Linwei Wang}
\affiliation{%
  \institution{Rochester Institute of Technology}
\streetaddress{54 Lomb Memorial Drive}
 \city{Rochester}
   \country{USA}}
\email{larst@affiliation.org}

\author{Jeff B. Pelz}
\affiliation{%
  \institution{Rochester Institute of Technology}
\streetaddress{54 Lomb Memorial Drive}
 \city{Rochester}
   \country{USA}}

\renewcommand{\shortauthors}{Chaudhary and Gyawali, et al.}

\begin{abstract}
Recent advances in appearance-based models have shown improved eye tracking performance in difficult scenarios like occlusion due to eyelashes, eyelids or camera placement, and environmental reflections on the cornea and glasses. The key reason for the improvement is the accurate and robust identification of eye parts (pupil, iris, and sclera regions). 
The improved accuracy often comes at the cost of 
labeling an enormous dataset, which is complex and time-consuming. 
This work presents two semi-supervised learning frameworks to identify eye-parts by taking advantage of unlabeled images where labeled datasets are scarce. With these frameworks, leveraging the domain-specific augmentation and novel spatially varying transformations for image segmentation, we show improved performance on various test cases. For instance, for a model trained on just 48 labeled images, these frameworks achieved an improvement of 0.38\% and 0.65\%  in segmentation performance over the baseline model, which is trained only with the labeled dataset.

\end{abstract}



\maketitle

\section{Introduction}

Effective gaze tracking can improve human-machine interactions by giving clues to users' behaviors and intentions and enhancing the experience for virtual, augmented, and mixed reality devices with efficient and realistic renderings by supporting foveated rendering and multi-focal displays to minimize vergence-accommodation conflicts~\cite{wu2019eyenet, garbin2019openeds}. Recent advances in gaze tracking using appearance-based gaze estimation~\cite{wu2019eyenet, palmero2020benefits, park2020towards} have shown robustness in person-independent scenarios, the effects of environmental reflections on the cornea and eyeglasses, occlusion due to eyelashes and camera placement, and heavy eye makeup. A backbone to most of the appearance-based models is the proper identification/segmentation of different parts of the eye. Recently, utilizing advancement in deep learning, ~\citet{yiu2019deepvog,chaudhary2019ritnet,kothari2020ellseg,wu2019eyenet} proposed different models to segment various eye parts accurately in challenging cases.

The success of these methods is often 
 predicated on the availability of large, curated training datasets of eye images with well-annotated labels. Such requirements are, however, difficult to satisfy in the general scenario. The acquisition process for the dataset for eye image segmentation requires a highly sophisticated and costly environment with significant (and often tedious) effort by experts. 
 Even in such a scenario, the dataset's quality may be limited by various factors like labeler's bias and inconsistency~\cite{kothari2020gaze}. As such, the acquisition process is difficult, if not impossible, for academic labs. Only a handful of 
 commercially funded labs have attempted to step in this direction of curating large datasets of eye image segmentation~\cite{garbin2019openeds, palmero2020openeds2020}. In this work, we utilize semi-supervised learning (SSL), which greatly diminishes the requirement for labeled data 
  by leveraging relatively easy-to-acquire unlabeled datasets.

SSL leverages unlabeled data to improve the learning from a small labeled dataset~\cite{zhou2003learning}. The primary goal of SSL algorithms is to avoid over-fitting of the model's parameters to a small amount of labeled data. Formally, in SSL, a data set $\mathcal{X} = \{ x_{1},x_{2},...,x_{n}\}$ is given among which only the first few $k$ 
 instances (images) are annotated $\{y_{1},...,y_{k}\} \in \mathcal{Y}$, and the remaining 
  instances are unlabeled. While learning the function $f: \mathcal{X}\rightarrow \mathcal{Y}$, 
SSL will exploit the hidden relationship within the data 
to predict the labels of unlabeled data points. One of the common inductive biases used to regularize SSL algorithms is the assumption of \textit{consistency} of the network function. Consistency refers to the fact that data points or their representations should have the same label predictions even after perturbation. Various deep learning-based work used the idea of consistency to perturb either data~\cite{laine2016temporal} or their hidden representations~\cite{gyawali2019semi}, and constrain the label predictions to be similar. 
These algorithms have demonstrated improved generalization performance in various domains, like image classification in computer vision and medical imaging. 
However, in semantic segmentation, this simple yet effective assumption of consistency is violated. In segmentation, output or label space accounts for the spatial position of pixels in the input space. Thus, even with small perceptual perturbation in the input space, we cannot enforce consistency in the output space. Thus, the recent progress made in the SSL literature has been mostly confined to the classification task, and very few works have explored SSL for semantic segmentation~\cite{ouali2020semi}. For the application of eye segmentation, the use of SSL is further limited.

This work first presents the SSL approach to utilize consistency training for semantic segmentation of eye images.
We use domain-specific augmentations that do not affect 
pixel positioning to perturb the input eye images and enforce consistency on the resulting model's predictions. Following this, we 
present a novel SSL method that uses an idea from self-supervised learning~\cite{kolesnikov2019revisiting} (formulate a new learning task) to strengthen the effect of regularization and improves generalization. This method also allows us to use commonly prescribed spatially varying augmentations like image rotation and translation while training the SSL method. We test these two SSL frameworks on one publicly available real eye segmentation dataset (OpenEDS-2019~\cite{garbin2019openeds}). We compare the presented methods' performance against the baseline, which includes training the model only with the labeled dataset. We further investigate the quality of the presented method for segmenting different eye parts, including iris and pupil. Finally, we present comparison studies on training the model with the labeled data from a single individual against a group of individuals on a fixed test set. In summary, the contributions of this work include:

\vspace{-3pt}
\begin{itemize}
    \item Adaptation of SSL setup for segmenting eye regions with domain-specific augmentations.
    \item A novel SSL framework to leverage spatially varying augmentations for semantic segmentation.
    \item Demonstration that a small number of 
    labeled images and a large number of 
     unlabeled images can significantly improve 
      eye image segmentation.
\end{itemize}

\section{Related Work}
A number of recent efforts have been made towards semantic segmentation of eye images to obtain the pupil, iris, and sclera regions~\cite{wu2019eyenet,kothari2020ellseg,chaudhary2019ritnet}.
\citet{wu2019eyenet} considers multiple heterogeneous tasks, including semantic segmentation, related to gaze estimation and propose an end-to-end deep learning method. Similarly, \citet{kothari2020ellseg} proposed a technique to improve pupil/iris center detection using an ellipse segmentation instead of segmenting the visible eye parts, and \citet{chaudhary2019ritnet} proposed a computationally efficient architecture to segment eye regions. Unlike those works that used large annotated image data sets to guide their learning process for segmentation of eye images, our work only considers a small amount of labeled data.

Semi-supervised learning (SSL) is one of the widely studied topics in machine learning. Consistency regularization is a commonly used regularization for SSL algorithms to exploit the hidden relationship between labeled and unlabeled data~\cite{zhou2003learning}. 
\citet{laine2016temporal} and \citet{gyawali2019semi} have used this idea of consistency in deep learning-based work and demonstrated improved performance in image classification tasks.
However, these works cannot be applied to the task of semantic segmentation as they enforce consistency on the output space by perturbing the pixels in the input space. 
In recent years, limited attempts have been made toward learning of semi-supervised semantic segmentation~\cite{ouali2020semi,kalluri2019universal}. For instance, \citet{kalluri2019universal} used pixel-level entropy regularization to train their semantic segmentation architecture. 
 While some other related work has explored SSL 
 for eye image segmentation, those studies
(e.g., \cite{shen2020domain}) considered extra information in the form of domain labels or considered a large number of labeled samples.

\begin{figure}
\begin{center}
\includegraphics[width=0.99\linewidth]{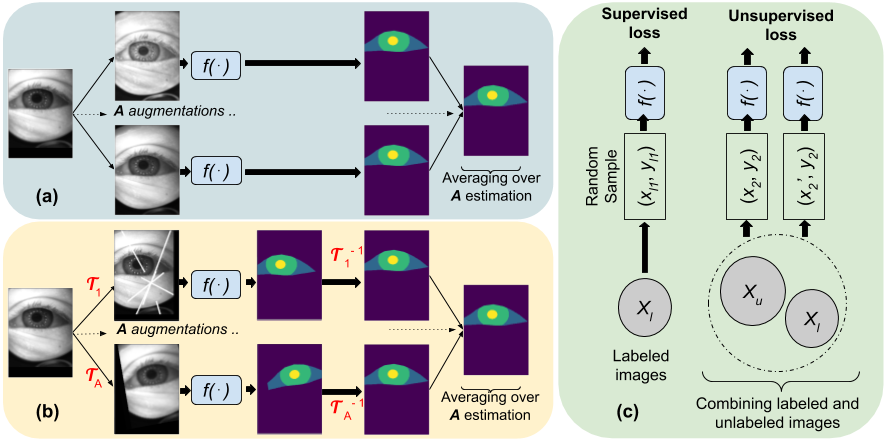}
\end{center}
\caption{Schematic diagram of the proposed SSL methods. For each unlabeled data, labels are guessed for $A$ separate copies generated via \textbf{(a)} SSL with domain-specific augmentation and \textbf{(b)} SSL with a self-supervised approach. In \textbf{(c)}, supervised loss and unsupervised loss are computed separately for labeled and a combination of labeled and unlabeled data set in both types of SSL methods.}
\label{fig:block}
\end{figure}

\section{Methods}
This section formulates the problem for the SSL framework. Following this we present two methods: SSL with domain-specific augmentation and a novel SSL with self-supervised learning.

\subsection{Problem formulation}
We consider a dataset with $k$ labeled training examples $\mathcal{X}_{l}$ with the corresponding labels $\mathcal{Y}_{l}$, and $m$ unlabeled training samples $\mathcal{X}_{u}$, where $k<m$.
Since our task is segmentation, the data space is represented by $\mathcal{X} \in \mathbb{R}^{CxHxW}$ where $HxW$ is the spatial dimension, and $C$ is the number of input channels. Similarly, the label space is represented by $\mathcal{Y} \in \mathbb{R}^{PxHxW}$ where $P$ is the number of classes.
We aim to learn parameters $\theta$ for the mapping function $f: \mathcal{X} \rightarrow\mathcal{Y}$, approximated via a deep neural network. 


\subsection{ $\mathcal{SSL}_{D}:$ SSL with domain-specific augmentation}
\label{sec:ssl}
Here, we present the SSL paradigm with domain-specific augmentation. To ensure effective utilization of consistency regularization, we employ augmentations without changing the image's spatial positions. Prior works for SSL utilize variation in noise perturbations~\cite{ouali2020semi}. In our observation, most of the eye images from the same hardware setup vary mostly in eye shape, skin/iris pigmentation, lighting conditions, gaze direction, eye with respect to camera, and blinks. Among them, variation on contrast/ luminance of eye images are referred to as domain-specific augmentation in this paper. Techniques like Contrast Limited Adaptive histogram equalization (CLAHE)~\cite{Zuiderveld1994ContrastEqualization} and Gamma correction have been shown advantageous for eye parts segmentation~\cite{chaudhary2019ritnet}. We leverage these domain-specific augmentations to perform label guessing in our SSL paradigm.

\subsubsection{Guessing Labels}
\label{sec:ssl_guess}
Using the domain augmentation strategies, we create $A$ separate copies for each data point and estimate their labels, as shown schematically 
in Fig. \ref{fig:block} (a). We compute the average of the model's prediction $y_u$ (softmax probabilities) for augmented copies of each data point $x_u$ as:
\begin{equation}
  \label{eq:labelEstimation}
y_u = \frac{1}{A}\sum_{a=1}^{A} f(x_{u,a}; \theta)
\end{equation}

Although the label guessing in this manner is commonly used for unlabeled datasets~\cite{berthelot2019mixmatch}, we find combining both labeled and unlabeled datasets to be beneficial. As such, in our case, any unlabeled data point $x_u$ is a random sample from the combination of both datasets $\mathcal{X}_{l}$ and $\mathcal{X}_{u}$. This way of guessing the label encourages the model to be consistent across different augmentations~\cite{gyawali2020semi}. Although following standard practice, we do not propagate gradients by computing the guessed labels; it should be noted that the guessed labels may change, as the segmentation network $f$ is updated over training. 

\subsubsection{SSL Objective}
\label{sec:ssl_obj}
We now have the guessed labels for $\mathcal{X}_{u}$ and true labels for $\mathcal{X}_{l}$. From $\mathcal{X}_{l}$, we randomly sample ($x_{l1}, y_{l1}$) as data-target pair to calculate supervised loss.
The objective for supervised setup is $\mathcal{L}_{sup} = \mathcal{L}_{CEL} (\lambda_1 + \lambda_2 \mathcal{L}_{BAL} ) + \lambda_3 \mathcal{L}_{SL}$, where  $\mathcal{L}_{CEL}$, $\mathcal{L}_{BAL}$, and $\mathcal{L}_{SL}$ are, respectively, the cross-entropy loss, boundary aware loss \cite{Ronneberger2015U-net:Segmentation} and the surface loss~\cite{Kervadec2018BoundarySegmentation}. For the case of $\mathcal{X}_{u}$, we consider unsupervised loss (referred to hereafter as $\mathcal{L}_{u}$), which is the L2 loss computed between the predicted softmax probabilities and the guessed label ($y_u$), as it is less susceptible to wrong predictions ~\cite{berthelot2019mixmatch,laine2016temporal}. 
As shown in Fig. \ref{fig:block} (c), we sample from the combination of labeled and unlabeled datasets, and consider all the augmented copies (Fig. \ref{fig:block} (c) represents the case when $A = 2$) to calculate $\mathcal{L}_{u}$. Since both data-point $x_{2}$, and $x_{2}^{'}$ have the same guesses label $y_{2}$, the minimization of unsupervised loss acts as the consistency regularization.  
The overall SSL loss is the combination of the $\mathcal{L}_{sup}$ and $\mathcal{L}_{u}$, with $\lambda_u$ as the weighting term between two-loss terms: $\mathcal{L} =\mathcal{L}_{sup} + \lambda_u \mathcal{L}_{u}$.




\subsection{ $\mathcal{SSL}_{ss}:$ SSL with self-supervised learning}
\label{sec:ssl_augu}
In this section, we propose a novel SSL method for the task of semantic segmentation utilizing the idea from \textit{self-supervised learning}~\cite{kolesnikov2019revisiting}. 
The self-supervised learning uses unlabeled data to formulate a pretext learning task such as predicting context, for which a target objective can be computed without supervision~\cite{kolesnikov2019revisiting}. In our case, predicting labels from the inversion of the transformed image's model prediction is the pretext learning task (Fig.~\ref{fig:block} (b)).
This also helps us to utilize widely used image transformations, including rotation and translation, which we were unable to use in Section \ref{sec:ssl}. Note that
these image transformations are important to account for variations in relative motion of the eye with respect to camera. 
We now present how we perform label guessing, which is the primary difference in comparison to $\mathcal{SSL}_{D}$. 

\subsubsection{Guessing Labels}
\label{sec:ssl_guess_augu}
We first consider the same domain augmentation strategies used in \ref{sec:ssl_guess} to create $A$ separate copies for each data point $x_u$. For each of 
these augmented data points, with different probability $p_{1}$, and $p_{2}$, we further introduce image transformation-based augmentation of rotation, and translation, respectively. Collectively, we will use $\mathcal{T}$ to denote these image transformations. We pass them into our segmentation network to obtain the corresponding labels and then calculate the inverse transform $\mathcal{T}^{-1}$ of these labels to bring them back into the same spatial space where the initial data points reside. Finally, we compute the average of the inverse of the model's prediction $y_u$, as the guessed labels for all $A$ separate copies of data point $x_u$: 

\begin{equation}
  \label{eq:labelaugu}
  y_u =\frac{1}{A}\sum_{a=1}^{A} \mathbf{\mathcal{T}^{-1}}(f(\mathbf{\mathcal{T}}(x_{u,a}); \theta))
\end{equation} 

This is represented schematically in Fig. \ref{fig:block} (b), and since the model has to be consistent on the predictions being invariant to both domain-specific augmentation and image transformations, this method of guessing the labels adds a more potent form of regularization compared to the one we presented in \ref{sec:ssl_guess}.

\subsubsection{SSL Objective}
Similar to \ref{sec:ssl_obj} and as shown schematically 
in Fig. \ref{fig:block} (c), we use the combination of supervised ($\mathcal{L}_{sup}$) and unsupervised ($\mathcal{L}_{u}$, $\mathcal{L}_{ss}$) loss for our objective function. $\mathcal{L}_{u}$ is the L2 loss computed between the predicted softmax prediction of data from domain-specific augmentation and the guessed label from \ref{sec:ssl_guess}. $\mathcal{L}_{ss}$ is the L2 loss computed between the same data and the guessed label from \ref{sec:ssl_guess_augu}. The overall objective function is $\mathcal{L} = \mathcal{L}_{sup} + \lambda_u \mathcal{L}_{u} + \lambda_{ss} \mathcal{L}_{ss}$, where $\lambda_u$ and $\lambda_{ss}$ are the corresponding weighting terms.


\section{Experiments}
\subsection{Datasets and Evaluation}
We evaluated our proposed pipeline on the OpenEDS-2019 dataset. OpenEDS-2019 has well-defined train ($n=8916$), validation (2403), and test (1440) sets where labels for the test set are not provided for public use. We evaluate our models on the validation set. Qualitative results on the test set is provided as supplementary video. Note that OpenEDS-2019 contains eye images of 94/28 (train/validation) different participants with multiple images. Further description of splits and test cases are explained in Section~\ref{results}. All images were resized to $240 \times 320$ for faster computation. For each dataset, models are tested on the fixed test set. Evaluations of the segmentation performance are with the standard mean Intersection over Union (IoU) score~\cite{everingham20052005}.

\subsection{Implementation Details}
As the primary purpose of the paper is designing an SSL paradigm to leverage the unlabeled dataset and not in designing a model architecture, we leverage the publicly available RITnet~\cite{chaudhary2019ritnet} that is computationally efficient. Models are trained using Adam optimizer ~\cite{Kingma2014Adam:Optimization} with a learning rate of 0.001 and a batch size of eight images (four labeled and four unlabeled) for 250 epochs. The model with the best validation score was reported. 

In our experiments, we used $\lambda_1=1$,  $\lambda_2=20$, $\lambda_3=1$. Two unsupervised loss hyper-parameters $\lambda_{u}$ and $\lambda_{ss}$ are two linearly increasing weights with slope 0.02 and 0.002 per epoch respectively. Initially, the loss scheduling scheme gives prominence to the supervised loss, and after a few epochs, unsupervised loss starts to increase.   

For domain-specific augmentation, we varied the contrast and luminance of the images. We selected random values in [0.8, 1.2] with step size of 0.05 for Gamma correction and random clip parameter in (1.0, 1.2, 1.5, 1.5, 1.5, 2.0) and grid size (2, 4, 8, 8, 8, 16) for CLAHE.  For $\mathcal{T}$ transform, random rotation [-5$^\circ$, 5$^\circ$] and translation [-20, 20 pixels] were performed with 50\% and 80\% probability respectively. Note the images in the dataset mostly varied in the translated form, so high importance was given to translation. Each image had 50\% chance of being transformed with $\mathcal{T}$. Additionally, we also used basic augmentations proposed in ~\cite{chaudhary2019ritnet} for all images.


\section{Results and Discussion}
\label{results}
The analysis of the results is based on two essential questions we want to address. First, does the availability of a large unlabeled dataset assist the learning from limited labeled data? Within this, we further ask what is the minimum number of labeled images per subject we can use in our pipeline to improve the performance significantly?
Second, will the model have better performance when the labeled images are from a single person, or when they are drawn from multiple subjects?
To answer these questions, we split our analysis into two parts: training with a varying number of labeled samples from multiple subjects and from a single subject. Further, we also compare the segmentation performance for essential eye features such as pupil and iris as it is important for reliable gaze estimation.




\subsection{Training with Multiple Subjects}
\label{sec:multiple}
\begin{figure}
\begin{center}
\includegraphics[width=0.7\linewidth]{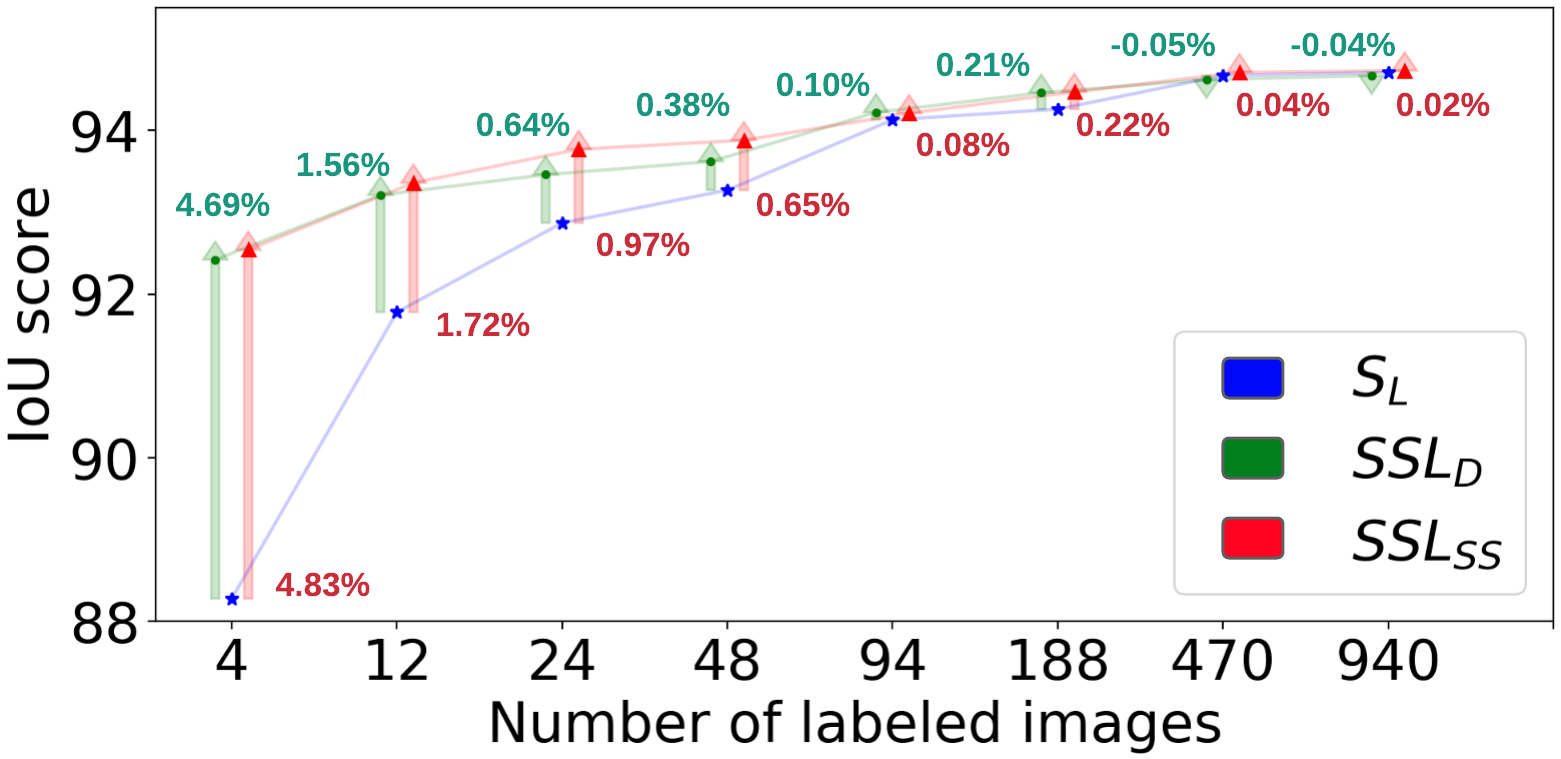}
\end{center}
\caption{IoU score for three cases ($\mathcal{S}_{L}$: blue, $\mathcal{SSL}_{D}$: green, and $\mathcal{SSL}_{SS}$: red) are shown with varying numbers of $\mathcal{X}_{l}$ and fixed $\mathcal{X}_{u}$. The number alongside arrows indicate respective improvement (in \%) over $\mathcal{S}_{L}$.}
\label{fig:multiple_subject}
\end{figure}

In this setup, the models were trained on varying numbers of labeled images ($\mathcal{X}_{l}$) and a fixed number of unlabeled images ($\mathcal{X}_{u}$ = 8916) from \textit{multiple subjects}. The models were evaluated on a fixed test set (2403). The equal number of images are selected from the same subject when $\mathcal{X}_{l}$ $\geq$ 94 and a random sample is chosen from different subjects for $\mathcal{X}_{l}$ $\leq$ 94. In Fig.~\ref{fig:multiple_subject}, the accuracy increases (88.28\% $\rightarrow$ 94.71\%) as $\mathcal{X}_{l}$ increases (4 $\rightarrow$ 940) when models are trained on labeled images only ($\mathcal{S}_{L}$). Similar behavior is observed with $\mathcal{SSL}_{D}$ (92.42\% $\rightarrow$ 94.67\%) framework. However, in all cases for $\mathcal{X}_{l}$$\leq$ 188, $\mathcal{SSL}_{D}$ achieves improved performance compared to training with $\mathcal{S}_{L}$ (e.g.,4.69\% for $\mathcal{X}_{l}$=4 \& 0.21\% for $\mathcal{X}_{l}$=94). This demonstrate that SSL framework can be helpful for eye image segmentation. Furthermore, the $\mathcal{SSL}_{SS}$ improves the performance against $\mathcal{S}_{L}$ 
(e.g.,4.83\% for $\mathcal{X}_{l}$=4 \& 0.02\% for $\mathcal{X}_{l}$=94)
and $\mathcal{SSL}_{D}$ by further upto 0.33\%. This improvement demonstrates effective utilization of unlabeled images. No clear benefits of SSL were observed for $\mathcal{X}_{l}$$\geq$ 188, which clearly signifies our proposed SSL frameworks are more suited when small amount of labeled data are available.


Most of the variation of model performance is expected in when number of $\mathcal{X}_{l}$ is small. To verify this, we trained our model by randomly selecting different subjects for cases when $\mathcal{X}_{l}$ is 4 and 12. The variations in the segmentation performance do not affect the incremental gain we observed from $\mathcal{S}_{L}$ to $\mathcal{SSL}_{D}$ and from $\mathcal{SSL}_{D}$ to $\mathcal{SSL}_{SS}$ (Table~\ref{tab:multile_subset_supplementaty}).

\begin{table}
\centering
\caption{Model performance for two subsets (along rows) when $\mathcal{X}_{l}$ is 4 \& 12 is shown for three frameworks ($\mathcal{S}_{L}$, $\mathcal{SSL}_{D}$, and $\mathcal{SSL}_{SS}$).
For each subset, \% change represents the improvement from $\mathcal{S}_{L}$ to $\mathcal{SSL}_{SS}$.}
\label{tab:multile_subset_supplementaty}       

\begin{tabular}{|l||l|l|l|l||l|l|l|l|l}
\cline{1-9}
\textbf{$\mathcal{X}_{l}$}   & \multicolumn{4}{l||}{4 images}                         & \multicolumn{4}{l|}{12 images}   &  \\ \cline{1-9}
\textit{Subset} & \multicolumn{1}{c|}{$\mathcal{S}_{L}$}    & $\mathcal{SSL}_{D}$ & $\mathcal{SSL}_{SS}$ & \% change & $\mathcal{S}_{L}$    & $\mathcal{SSL}_{D}$  & $\mathcal{SSL}_{SS}$  & \% change &  \\ \cline{1-9}
I      & \multicolumn{1}{c|}{87.42} & 91.59 & 92.04 & \textbf{5.28}     & 91.78 & 93.21 & 93.36 & \textbf{1.72}     &  \\ \cline{1-9}
II     & 88.28                      & 92.42 & 92.54 & \textbf{4.83}     & 91.52 & 92.81 & 93.05 & \textbf{1.67}     &  \\ \cline{1-9}
\end{tabular}
\end{table}

\subsection{Training with a Single Subject}
\label{sec:single}
\begin{figure}
\begin{center}
\includegraphics[width=0.9\linewidth]{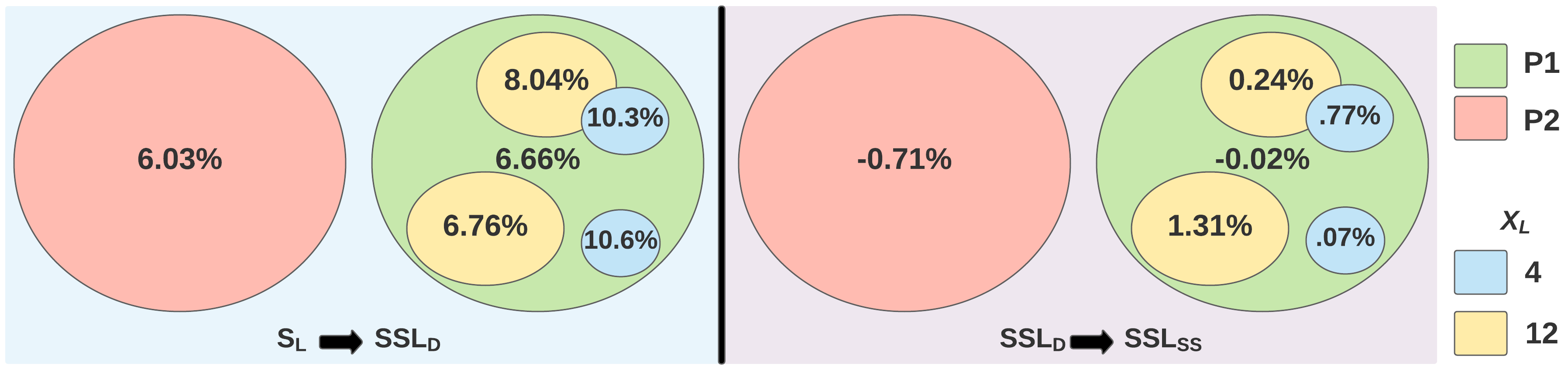}
\end{center}
\caption{Demonstration of improvement (in \%) for cases  $\mathcal{S}_{L}$ to $\mathcal{SSL}_{D}$ (left) and  $\mathcal{SSL}_{D}$ to $\mathcal{SSL}_{SS}$ (right) when models are trained on two subjects (red and green). For P1 (green), we further test the change in model performance for various subsets of $\mathcal{X}_{l}$.}
\label{fig:single_subject}
\end{figure}

Here, we consider the effect of training the model using $\mathcal{X}_{l}$ from a \textit{single subject} and a fixed $\mathcal{X}_{u}$ from multiple subjects. For each dataset, we randomly selected two subjects (P1 and P2) for our analysis. For each subject, we use all the available samples as $\mathcal{X}_{l}$  to train out model. The results are demonstrated in Fig.~\ref{fig:single_subject} (left and right), respectively, as percentage improvement as we go from $\mathcal{S}_L$ to $\mathcal{SSL}_{D}$ and from $\mathcal{SSL}_{D}$ to $\mathcal{SSL}_{SS}$. For P1, we further experiment using varying subsets of $\mathcal{X}_{l}$ (4 and 12) from a single subject. 
Note that the addition of $\mathcal{SSL}_{SS}$ provided additional boost up to 1.31\% compared to $\mathcal{SSL}_{D}$. The only cases where $\mathcal{SSL}_{SS}$ did not outperform $\mathcal{SSL}_{D}$ was when trained on large samples of images of a single person (as shown in Fig. ~\ref{fig:single_subject}).

Compared to (Section \ref{sec:multiple}), where we trained with multiple subjects, training with a single subject degrades the performance severely for $\mathcal{S}_L$ (e.g. 88.28 \% $\rightarrow$ 83.32\% when trained on $\mathcal{X}_{l} = 4$). We believe inherent variations among subjects helped while training with multiple subjects. However, in both SSL frameworks, with the addition of $\mathcal{X}_{u}$, this gap is reduced  (e.g. 92.54 \% $\rightarrow$ 92.20\% when trained on $\mathcal{X}_{l} = 4$ for $\mathcal{SSL}_{SS}$). 

\subsection{Eye part segmentation}
\begin{figure}
\begin{center}
\includegraphics[width=0.8\linewidth]{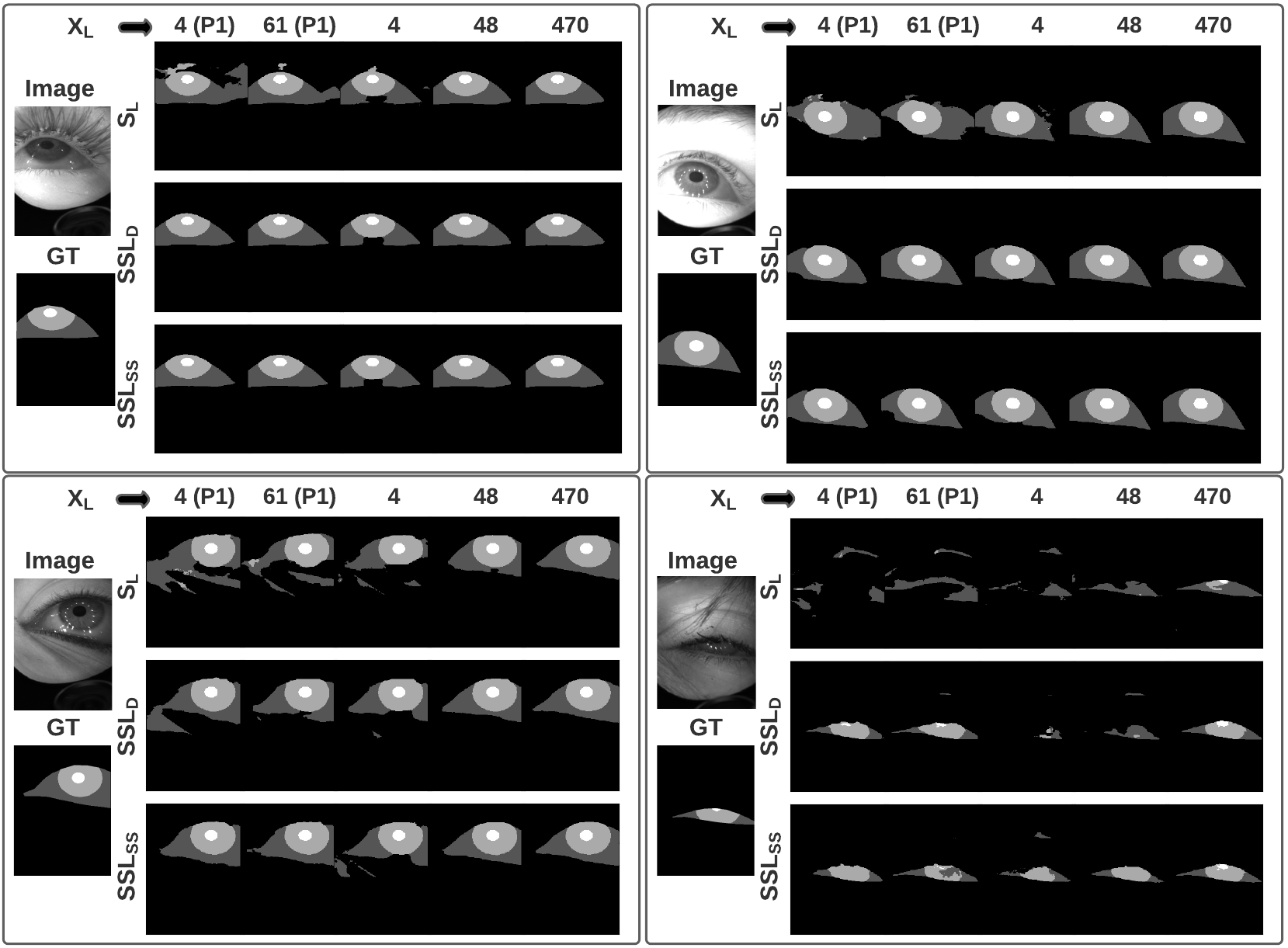}
\end{center}
\caption{Four samples of the test set with its corresponding ground-truth and network predictions for the number of cases are shown in adjacent blocks. As the number of images increases, the confidence in prediction and unwanted spurious patches are reduced when models are trained on $\mathcal{S}_{L}$. For SSL approaches, the confidence is prediction is more even when a small number of $\mathcal{X}_{l}$ are used. No significant difference is visible for the two SSL approaches, which mostly vary in finer details.}
\label{fig:samples}
\end{figure}

In Table ~\ref{tab:pupil}, we present the IoU score separately for the pupil and iris classes. 
Similar behavior is observed as Section ~\ref{sec:multiple} and ~\ref{sec:single}, suggesting that every class's accuracy is improved with these frameworks. Note that higher segmentation accuracy means a better chance of ellipse fits and thus accurate gaze estimation. 
In Fig. ~\ref{fig:samples}, we present a predicted segmentation mask for various test cases for qualitative comparison. Here, the spurious patches start to disappear as the number of labeled images are increased or with the introduction of unlabeled images with the SSL framework.  

\begin{table}[]

\caption{
Comparison of pupil and iris (inside parenthesis) class IoU scores for three frameworks ($\mathcal{S}_{L}$, $\mathcal{SSL}_{D}$, and $\mathcal{SSL}_{SS}$) for varying number of $\mathcal{X}_{l}$ and fixed number of $\mathcal{X}_{u}$. P1 indicates samples from a single subject. Bold values indicate the best performance within each test case (along column). \% change represents the improvement from $\mathcal{S}_{L}$ to $\mathcal{SSL}_{SS}$}
\label{tab:pupil}       
\begin{adjustbox}{width=\columnwidth,center}
\begin{tabular}{|c|ccccccc|}
\hline
$\mathcal{X}_{l}$     & 4                      & 12                     & 24                     & 48                     & 4 (P1)                 & 12 (P1)                & 61 (P1)                \\ \hline
{$\mathcal{S}_{L}$}     & 88.25 (89.06)          & 91.84 (92.87)          & 92.31 (93.25)          & 92.60 (93.55)          & 88.00 (86.98)          & 87.83 (88.22)          & 90.92 (90.24)          \\ \hline
{$\mathcal{SSL}_{D}$}   & \textbf{91.97} (92.75) & 92.50 (94.00)          & 92.69 (94.01)          & 92.69 (94.02)          & 91.73 (92.84)          & 91.63 (93.17)          & 92.67 (93.77)          \\ \hline
{$\mathcal{SSL}_{SS}$} & 91.85 \textbf{(93.05)} & \textbf{92.62 (94.14)} & \textbf{92.81 (94.17)} & \textbf{93.01 (94.28)} & \textbf{91.87 (93.35)} & \textbf{91.78 (93.62)} & \textbf{92.86 (93.93)} \\ \hline\hline
\% change   & 4.08 (4.48)            & 0.85 (1.37)            & 0.54 (0.99)            & 0.44 (0.78)            & 4.4.40 (7.32)          & 4.50 (6.12)            & 2.13 (4.09)            \\ \hline
\end{tabular}
\end{adjustbox}
\end{table}

\section{Conclusion and Future Work}




We present two SSL frameworks that leverage the domain-specific augmentations and pretext learning task that accounts for spatially varying transformations. With this, we demonstrate a substantial increase in segmentation performance for a small number of labeled images by considering hidden relationships present in a large number of unlabeled images. The efficacy of the two frameworks is demonstrated on OpenEDS-2019 datasets. This paper opens an exciting area for the eye-tracking community to focus on the variability of the subjects rather than labeling a large number of images of a particular subject. 
In the future, we would like to investigate the effect of curating labeled datasets (e.g., considering eye position and blinks) instead of random selection in the SSL framework. Pre-trained models and source code will be made available.
\bibliographystyle{ACM-Reference-Format}
\bibliography{ref}


\begin{thebibliography}{23}


\ifx \showCODEN    \undefined \def \showCODEN     #1{\unskip}     \fi
\ifx \showDOI      \undefined \def \showDOI       #1{#1}\fi
\ifx \showISBNx    \undefined \def \showISBNx     #1{\unskip}     \fi
\ifx \showISBNxiii \undefined \def \showISBNxiii  #1{\unskip}     \fi
\ifx \showISSN     \undefined \def \showISSN      #1{\unskip}     \fi
\ifx \showLCCN     \undefined \def \showLCCN      #1{\unskip}     \fi
\ifx \shownote     \undefined \def \shownote      #1{#1}          \fi
\ifx \showarticletitle \undefined \def \showarticletitle #1{#1}   \fi
\ifx \showURL      \undefined \def \showURL       {\relax}        \fi
\providecommand\bibfield[2]{#2}
\providecommand\bibinfo[2]{#2}
\providecommand\natexlab[1]{#1}
\providecommand\showeprint[2][]{arXiv:#2}

\bibitem[\protect\citeauthoryear{Berthelot, Carlini, Goodfellow, Papernot,
  Oliver, and Raffel}{Berthelot et~al\mbox{.}}{2019}]%
        {berthelot2019mixmatch}
\bibfield{author}{\bibinfo{person}{David Berthelot}, \bibinfo{person}{Nicholas
  Carlini}, \bibinfo{person}{Ian Goodfellow}, \bibinfo{person}{Nicolas
  Papernot}, \bibinfo{person}{Avital Oliver}, {and} \bibinfo{person}{Colin~A
  Raffel}.} \bibinfo{year}{2019}\natexlab{}.
\newblock \showarticletitle{Mixmatch: A holistic approach to semi-supervised
  learning}. In \bibinfo{booktitle}{\emph{Advances in Neural Information
  Processing Systems}}. \bibinfo{pages}{5049--5059}.
\newblock


\bibitem[\protect\citeauthoryear{Chaudhary, Kothari, Acharya, Dangi, Nair,
  Bailey, Kanan, Diaz, and Pelz}{Chaudhary et~al\mbox{.}}{2019}]%
        {chaudhary2019ritnet}
\bibfield{author}{\bibinfo{person}{Aayush~K Chaudhary},
  \bibinfo{person}{Rakshit Kothari}, \bibinfo{person}{Manoj Acharya},
  \bibinfo{person}{Shusil Dangi}, \bibinfo{person}{Nitinraj Nair},
  \bibinfo{person}{Reynold Bailey}, \bibinfo{person}{Christopher Kanan},
  \bibinfo{person}{Gabriel Diaz}, {and} \bibinfo{person}{Jeff~B Pelz}.}
  \bibinfo{year}{2019}\natexlab{}.
\newblock \showarticletitle{RITnet: real-time semantic segmentation of the eye
  for gaze tracking}. In \bibinfo{booktitle}{\emph{2019 IEEE/CVF International
  Conference on Computer Vision Workshop (ICCVW)}}. IEEE,
  \bibinfo{pages}{3698--3702}.
\newblock


\bibitem[\protect\citeauthoryear{Everingham, Zisserman, Williams, Van~Gool,
  Allan, Bishop, Chapelle, Dalal, Deselaers, Dork{\'o},
  et~al\mbox{.}}{Everingham et~al\mbox{.}}{2005}]%
        {everingham20052005}
\bibfield{author}{\bibinfo{person}{Mark Everingham}, \bibinfo{person}{Andrew
  Zisserman}, \bibinfo{person}{Christopher~KI Williams}, \bibinfo{person}{Luc
  Van~Gool}, \bibinfo{person}{Moray Allan}, \bibinfo{person}{Christopher~M
  Bishop}, \bibinfo{person}{Olivier Chapelle}, \bibinfo{person}{Navneet Dalal},
  \bibinfo{person}{Thomas Deselaers}, \bibinfo{person}{Gyuri Dork{\'o}},
  {et~al\mbox{.}}} \bibinfo{year}{2005}\natexlab{}.
\newblock \showarticletitle{The 2005 pascal visual object classes challenge}.
  In \bibinfo{booktitle}{\emph{Machine Learning Challenges Workshop}}.
  Springer, \bibinfo{pages}{117--176}.
\newblock


\bibitem[\protect\citeauthoryear{Garbin, Shen, Schuetz, Cavin, Hughes, and
  Talathi}{Garbin et~al\mbox{.}}{2019}]%
        {garbin2019openeds}
\bibfield{author}{\bibinfo{person}{Stephan~J Garbin}, \bibinfo{person}{Yiru
  Shen}, \bibinfo{person}{Immo Schuetz}, \bibinfo{person}{Robert Cavin},
  \bibinfo{person}{Gregory Hughes}, {and} \bibinfo{person}{Sachin~S Talathi}.}
  \bibinfo{year}{2019}\natexlab{}.
\newblock \showarticletitle{Openeds: Open eye dataset}.
\newblock \bibinfo{journal}{\emph{arXiv preprint arXiv:1905.03702}}
  (\bibinfo{year}{2019}).
\newblock


\bibitem[\protect\citeauthoryear{Gyawali, Ghimire, Bajracharya, Li, and
  Wang}{Gyawali et~al\mbox{.}}{2020}]%
        {gyawali2020semi}
\bibfield{author}{\bibinfo{person}{Prashnna~Kumar Gyawali},
  \bibinfo{person}{Sandesh Ghimire}, \bibinfo{person}{Pradeep Bajracharya},
  \bibinfo{person}{Zhiyuan Li}, {and} \bibinfo{person}{Linwei Wang}.}
  \bibinfo{year}{2020}\natexlab{}.
\newblock \showarticletitle{Semi-supervised Medical Image Classification with
  Global Latent Mixing}.
\newblock \bibinfo{journal}{\emph{arXiv preprint arXiv:2005.11217}}
  (\bibinfo{year}{2020}).
\newblock


\bibitem[\protect\citeauthoryear{Gyawali, Li, Ghimire, and Wang}{Gyawali
  et~al\mbox{.}}{2019}]%
        {gyawali2019semi}
\bibfield{author}{\bibinfo{person}{Prashnna~Kumar Gyawali},
  \bibinfo{person}{Zhiyuan Li}, \bibinfo{person}{Sandesh Ghimire}, {and}
  \bibinfo{person}{Linwei Wang}.} \bibinfo{year}{2019}\natexlab{}.
\newblock \showarticletitle{Semi-supervised learning by disentangling and
  self-ensembling over stochastic latent space}. In
  \bibinfo{booktitle}{\emph{International Conference on Medical Image Computing
  and Computer-Assisted Intervention}}. Springer, \bibinfo{pages}{766--774}.
\newblock


\bibitem[\protect\citeauthoryear{Kalluri, Varma, Chandraker, and
  Jawahar}{Kalluri et~al\mbox{.}}{2019}]%
        {kalluri2019universal}
\bibfield{author}{\bibinfo{person}{Tarun Kalluri}, \bibinfo{person}{Girish
  Varma}, \bibinfo{person}{Manmohan Chandraker}, {and} \bibinfo{person}{CV
  Jawahar}.} \bibinfo{year}{2019}\natexlab{}.
\newblock \showarticletitle{Universal semi-supervised semantic segmentation}.
  In \bibinfo{booktitle}{\emph{Proceedings of the IEEE International Conference
  on Computer Vision}}. \bibinfo{pages}{5259--5270}.
\newblock


\bibitem[\protect\citeauthoryear{Kervadec, Bouchtiba, Desrosiers, Granger,
  Dolz, and Ayed}{Kervadec et~al\mbox{.}}{2018}]%
        {Kervadec2018BoundarySegmentation}
\bibfield{author}{\bibinfo{person}{Hoel Kervadec}, \bibinfo{person}{Jihene
  Bouchtiba}, \bibinfo{person}{Christian Desrosiers}, \bibinfo{person}{Éric
  Granger}, \bibinfo{person}{Jose Dolz}, {and} \bibinfo{person}{Ismail~Ben
  Ayed}.} \bibinfo{year}{2018}\natexlab{}.
\newblock \bibinfo{title}{{Boundary loss for highly unbalanced segmentation}}.
\newblock
\newblock
\urldef\tempurl%
\url{http://arxiv.org/abs/1812.07032}
\showURL{%
\tempurl}


\bibitem[\protect\citeauthoryear{Kingma and Ba}{Kingma and Ba}{2014}]%
        {Kingma2014Adam:Optimization}
\bibfield{author}{\bibinfo{person}{Diederik~P. Kingma} {and}
  \bibinfo{person}{Jimmy Ba}.} \bibinfo{year}{2014}\natexlab{}.
\newblock \showarticletitle{{Adam: A Method for Stochastic Optimization}}.
\newblock \bibinfo{journal}{\emph{Journal of neuroscience methods}}
  \bibinfo{volume}{148}, \bibinfo{number}{2} (\bibinfo{date}{12}
  \bibinfo{year}{2014}), \bibinfo{pages}{167--76}.
\newblock
\showISBNx{0925-2312 1872-8286}
\showISSN{0165-0270}
\urldef\tempurl%
\url{https://doi.org/10.1016/j.jneumeth.2005.04.009}
\showDOI{\tempurl}


\bibitem[\protect\citeauthoryear{Kolesnikov, Zhai, and Beyer}{Kolesnikov
  et~al\mbox{.}}{2019}]%
        {kolesnikov2019revisiting}
\bibfield{author}{\bibinfo{person}{Alexander Kolesnikov},
  \bibinfo{person}{Xiaohua Zhai}, {and} \bibinfo{person}{Lucas Beyer}.}
  \bibinfo{year}{2019}\natexlab{}.
\newblock \showarticletitle{Revisiting self-supervised visual representation
  learning}. In \bibinfo{booktitle}{\emph{Proceedings of the IEEE conference on
  Computer Vision and Pattern Recognition}}. \bibinfo{pages}{1920--1929}.
\newblock


\bibitem[\protect\citeauthoryear{Kothari, Yang, Kanan, Bailey, Pelz, and
  Diaz}{Kothari et~al\mbox{.}}{2020b}]%
        {kothari2020gaze}
\bibfield{author}{\bibinfo{person}{Rakshit Kothari}, \bibinfo{person}{Zhizhuo
  Yang}, \bibinfo{person}{Christopher Kanan}, \bibinfo{person}{Reynold Bailey},
  \bibinfo{person}{Jeff~B Pelz}, {and} \bibinfo{person}{Gabriel~J Diaz}.}
  \bibinfo{year}{2020}\natexlab{b}.
\newblock \showarticletitle{Gaze-in-wild: A dataset for studying eye and head
  coordination in everyday activities}.
\newblock \bibinfo{journal}{\emph{Scientific reports}} \bibinfo{volume}{10},
  \bibinfo{number}{1} (\bibinfo{year}{2020}), \bibinfo{pages}{1--18}.
\newblock


\bibitem[\protect\citeauthoryear{Kothari, Chaudhary, Bailey, Pelz, and
  Diaz}{Kothari et~al\mbox{.}}{2020a}]%
        {kothari2020ellseg}
\bibfield{author}{\bibinfo{person}{Rakshit~S Kothari},
  \bibinfo{person}{Aayush~K Chaudhary}, \bibinfo{person}{Reynold~J Bailey},
  \bibinfo{person}{Jeff~B Pelz}, {and} \bibinfo{person}{Gabriel~J Diaz}.}
  \bibinfo{year}{2020}\natexlab{a}.
\newblock \showarticletitle{EllSeg: An Ellipse Segmentation Framework for
  Robust Gaze Tracking}.
\newblock \bibinfo{journal}{\emph{arXiv preprint arXiv:2007.09600}}
  (\bibinfo{year}{2020}).
\newblock


\bibitem[\protect\citeauthoryear{Laine and Aila}{Laine and Aila}{2017}]%
        {laine2016temporal}
\bibfield{author}{\bibinfo{person}{Samuli Laine} {and} \bibinfo{person}{Timo
  Aila}.} \bibinfo{year}{2017}\natexlab{}.
\newblock \showarticletitle{Temporal ensembling for semi-supervised learning}.
  In \bibinfo{booktitle}{\emph{ICLR}}.
\newblock


\bibitem[\protect\citeauthoryear{Ouali, Hudelot, and Tami}{Ouali
  et~al\mbox{.}}{2020}]%
        {ouali2020semi}
\bibfield{author}{\bibinfo{person}{Yassine Ouali}, \bibinfo{person}{C{\'e}line
  Hudelot}, {and} \bibinfo{person}{Myriam Tami}.}
  \bibinfo{year}{2020}\natexlab{}.
\newblock \showarticletitle{Semi-Supervised Semantic Segmentation with
  Cross-Consistency Training}. In \bibinfo{booktitle}{\emph{Proceedings of the
  IEEE/CVF Conference on Computer Vision and Pattern Recognition}}.
  \bibinfo{pages}{12674--12684}.
\newblock


\bibitem[\protect\citeauthoryear{Palmero, Komogortsev, and Talathi}{Palmero
  et~al\mbox{.}}{2020a}]%
        {palmero2020benefits}
\bibfield{author}{\bibinfo{person}{Cristina Palmero}, \bibinfo{person}{Oleg~V
  Komogortsev}, {and} \bibinfo{person}{Sachin~S Talathi}.}
  \bibinfo{year}{2020}\natexlab{a}.
\newblock \showarticletitle{Benefits of temporal information for
  appearance-based gaze estimation}.
\newblock \bibinfo{journal}{\emph{arXiv preprint arXiv:2005.11670}}
  (\bibinfo{year}{2020}).
\newblock


\bibitem[\protect\citeauthoryear{Palmero, Sharma, Behrendt, Krishnakumar,
  Komogortsev, and Talathi}{Palmero et~al\mbox{.}}{2020b}]%
        {palmero2020openeds2020}
\bibfield{author}{\bibinfo{person}{Cristina Palmero}, \bibinfo{person}{Abhishek
  Sharma}, \bibinfo{person}{Karsten Behrendt}, \bibinfo{person}{Kapil
  Krishnakumar}, \bibinfo{person}{Oleg~V Komogortsev}, {and}
  \bibinfo{person}{Sachin~S Talathi}.} \bibinfo{year}{2020}\natexlab{b}.
\newblock \showarticletitle{OpenEDS2020: Open Eyes Dataset}.
\newblock \bibinfo{journal}{\emph{arXiv preprint arXiv:2005.03876}}
  (\bibinfo{year}{2020}).
\newblock


\bibitem[\protect\citeauthoryear{Park, Aksan, Zhang, and Hilliges}{Park
  et~al\mbox{.}}{2020}]%
        {park2020towards}
\bibfield{author}{\bibinfo{person}{Seonwook Park}, \bibinfo{person}{Emre
  Aksan}, \bibinfo{person}{Xucong Zhang}, {and} \bibinfo{person}{Otmar
  Hilliges}.} \bibinfo{year}{2020}\natexlab{}.
\newblock \showarticletitle{Towards End-to-end Video-based Eye-Tracking}. In
  \bibinfo{booktitle}{\emph{European Conference on Computer Vision}}. Springer,
  \bibinfo{pages}{747--763}.
\newblock


\bibitem[\protect\citeauthoryear{Ronneberger, Fischer, and Brox}{Ronneberger
  et~al\mbox{.}}{2015}]%
        {Ronneberger2015U-net:Segmentation}
\bibfield{author}{\bibinfo{person}{Olaf Ronneberger}, \bibinfo{person}{Philipp
  Fischer}, {and} \bibinfo{person}{Thomas Brox}.}
  \bibinfo{year}{2015}\natexlab{}.
\newblock \showarticletitle{{U-net: Convolutional networks for biomedical image
  segmentation}}.
\newblock \bibinfo{journal}{\emph{International Conference on Medical image
  computing and computer-assisted intervention. Springer, Cham, 2015.}}
  \bibinfo{volume}{9351} (\bibinfo{year}{2015}), \bibinfo{pages}{234--241}.
\newblock
\showISBNx{9783319245737}
\showISSN{16113349}
\urldef\tempurl%
\url{https://doi.org/10.1007/978-3-319-24574-4{\_}28}
\showDOI{\tempurl}


\bibitem[\protect\citeauthoryear{Shen, Komogortsev, and Talathi}{Shen
  et~al\mbox{.}}{2020}]%
        {shen2020domain}
\bibfield{author}{\bibinfo{person}{Yiru Shen}, \bibinfo{person}{Oleg
  Komogortsev}, {and} \bibinfo{person}{Sachin~S Talathi}.}
  \bibinfo{year}{2020}\natexlab{}.
\newblock \showarticletitle{Domain Adaptation for Eye Segmentation}. In
  \bibinfo{booktitle}{\emph{European Conference on Computer Vision}}. Springer,
  \bibinfo{pages}{555--569}.
\newblock


\bibitem[\protect\citeauthoryear{Wu, Rajendran, Van~As, Badrinarayanan, and
  Rabinovich}{Wu et~al\mbox{.}}{2019}]%
        {wu2019eyenet}
\bibfield{author}{\bibinfo{person}{Zhengyang Wu}, \bibinfo{person}{Srivignesh
  Rajendran}, \bibinfo{person}{Tarrence Van~As}, \bibinfo{person}{Vijay
  Badrinarayanan}, {and} \bibinfo{person}{Andrew Rabinovich}.}
  \bibinfo{year}{2019}\natexlab{}.
\newblock \showarticletitle{EyeNet: A Multi-Task Deep Network for Off-Axis Eye
  Gaze Estimation}. In \bibinfo{booktitle}{\emph{2019 IEEE/CVF International
  Conference on Computer Vision Workshop (ICCVW)}}. IEEE,
  \bibinfo{pages}{3683--3687}.
\newblock


\bibitem[\protect\citeauthoryear{Yiu, Aboulatta, Raiser, Ophey, Flanagin,
  zu~Eulenburg, and Ahmadi}{Yiu et~al\mbox{.}}{2019}]%
        {yiu2019deepvog}
\bibfield{author}{\bibinfo{person}{Yuk-Hoi Yiu}, \bibinfo{person}{Moustafa
  Aboulatta}, \bibinfo{person}{Theresa Raiser}, \bibinfo{person}{Leoni Ophey},
  \bibinfo{person}{Virginia~L Flanagin}, \bibinfo{person}{Peter zu Eulenburg},
  {and} \bibinfo{person}{Seyed-Ahmad Ahmadi}.} \bibinfo{year}{2019}\natexlab{}.
\newblock \showarticletitle{DeepVOG: Open-source pupil segmentation and gaze
  estimation in neuroscience using deep learning}.
\newblock \bibinfo{journal}{\emph{Journal of neuroscience methods}}
  \bibinfo{volume}{324} (\bibinfo{year}{2019}), \bibinfo{pages}{108307}.
\newblock


\bibitem[\protect\citeauthoryear{Zhou, Bousquet, Lal, Weston, and
  Sch{\"o}lkopf}{Zhou et~al\mbox{.}}{2003}]%
        {zhou2003learning}
\bibfield{author}{\bibinfo{person}{Dengyong Zhou}, \bibinfo{person}{Olivier
  Bousquet}, \bibinfo{person}{Thomas Lal}, \bibinfo{person}{Jason Weston},
  {and} \bibinfo{person}{Bernhard Sch{\"o}lkopf}.}
  \bibinfo{year}{2003}\natexlab{}.
\newblock \showarticletitle{Learning with local and global consistency}.
\newblock \bibinfo{journal}{\emph{Advances in neural information processing
  systems}}  \bibinfo{volume}{16} (\bibinfo{year}{2003}),
  \bibinfo{pages}{321--328}.
\newblock


\bibitem[\protect\citeauthoryear{Zuiderveld}{Zuiderveld}{1994}]%
        {Zuiderveld1994ContrastEqualization}
\bibfield{author}{\bibinfo{person}{Karel Zuiderveld}.}
  \bibinfo{year}{1994}\natexlab{}.
\newblock \showarticletitle{{Contrast Limited Adaptive Histogram
  Equalization}}.
\newblock In \bibinfo{booktitle}{\emph{Graphics Gems}}.
\newblock
\urldef\tempurl%
\url{https://doi.org/10.1016/b978-0-12-336156-1.50061-6}
\showDOI{\tempurl}


\end{thebibliography}

\end{document}